%% file: main.tex
\documentclass[pmlr]{jmlr}

\usepackage{booktabs}
\usepackage{multirow}
\usepackage{adjustbox}
\usepackage{array}
\usepackage{amsmath,amssymb,graphicx,url}
\usepackage{enumitem}
\usepackage{pifont}
\usepackage{booktabs,multirow,array,makecell}
\usepackage{makecell}

\jmlrvolume{329}
\jmlryear{2026}
\jmlrworkshop{Conformal and Probabilistic Prediction with Applications}
\jmlrproceedings{PMLR}{Proceedings of Machine Learning Research}

\title[A Deployment Audit of Release-Side Risk in Conformal Triage under Prevalence Shift]{A Deployment Audit of Release-Side Risk in Conformal Triage under Prevalence Shift}

\author{\Name{Chengze Li} \Email{chengze6@uic.edu}\\
\addr University of Illinois Chicago, Chicago, IL, USA
\AND
\Name{Xiao Liu} \Email{liuxiao@manteiatech.com}\\
\addr Manteia Technologies Co., Ltd, Xiamen, China
\AND
\Name{Hanrong Zhang} \Email{hzhan135@uic.edu}\\
\addr University of Illinois Chicago, Chicago, IL, USA
\AND
\Name{Haiyang Peng} \Email{penghaiyang@manteiatech.com}\\
\addr Manteia Technologies Co., Ltd, Xiamen, China
\AND
\Name{Yanghao Ruan} \Email{yanghao4@illinois.edu}\\
\addr University of Illinois Urbana-Champaign, Champaign, IL, USA
\AND
\Name{Huanhuan Ma} \Email{hma42@uic.edu}\\
\addr University of Illinois Chicago, Chicago, IL, USA
\AND
\Name{Chunyu Miao} \Email{cmiao8@uic.edu}\\
\addr University of Illinois Chicago, Chicago, IL, USA
\AND
\Name{Qichao Zhou} \Email{zhouqc@manteiatech.com}\\
\addr Manteia Technologies Co., Ltd, Xiamen, China
\AND
\Name{Xiangrong Qi} \Email{xqi@mednet.ucla.edu}\\
\addr University of California Los Angeles, Los Angeles, CA, USA
\AND
\Name{Philip Yu} \Email{psyu@uic.edu}\\
\addr University of Illinois Chicago, Chicago, IL, USA}

\editor{Ernst Ahlberg, Ulf Johansson, Henrik Bostr\"om,
Alberto Carlevaro, Johan Hallberg \mbox{Szabadv\'ary}
and Lars Carlsson}

\newcommand{\Cmarg}{C_{\mathrm{marg}}}
\newcommand{\Cev}{C_{\mathrm{ev}}}
\newcommand{\FNrel}{\mathrm{FN}_{\mathrm{rel}}}
\newcommand{\FNp}{Q_{0.95}(\FNrel)}
\newcommand{\Prel}{P_{\mathrm{rel}}}
\newcommand{\HRR}{\mathrm{HRR}}

\newcommand{\sd}[1]{\,{\scriptstyle\pm #1}}
\renewcommand{\sd}[1]{\,\pm #1}

\begin{document}
\maketitle

\begin{abstract}
Conformal triage converts predictive scores into deployment actions that either release a case, flag it for urgent attention, or defer it to human review. Under an observed change in target-event prevalence, however, marginal coverage and human-review rate can miss whether patients who experience the target event are released without review. To address this gap, we introduce a leakage-aware deployment audit for release-side conformal triage. It first assigns target subjects to three non-overlapping roles: prevalence correction, conformal calibration, and held-out release-side evaluation. This separation then lets the audit evaluate release directly: how many event-positive patients are cleared without review, whether the pilot has enough event labels for calibration, and how the release-review trade-off shifts. Applying this audit to a retrospective non-small-cell lung cancer (NSCLC) target cohort shows why lower review can be misleading: after prevalence correction, the pooled conformal branch lowers review by releasing more patients, some of whom are event-positive. Within the audit, the classwise branch acts as a scarcity diagnostic: the pilot has too few event labels to support a low-review release rule.
\end{abstract}


\section{Introduction}
\label{sec:intro}

\begin{figure}[!t]
\floatconts
  {fig:concept}
  {\caption{Release-side risk in pooled conformal triage. (A) Conformal triage maps each score to release, flag, or defer. (B) Pooled split conformal prediction sets one threshold using all calibration patients; when the target event prevalence is about 23\%, the threshold is dominated by non-event patients. (C) A monotone prevalence correction can move the pooled release boundary while marginal coverage remains stable, allowing true event cases to be released without review.}}
  {\includegraphics[width=0.95\linewidth]{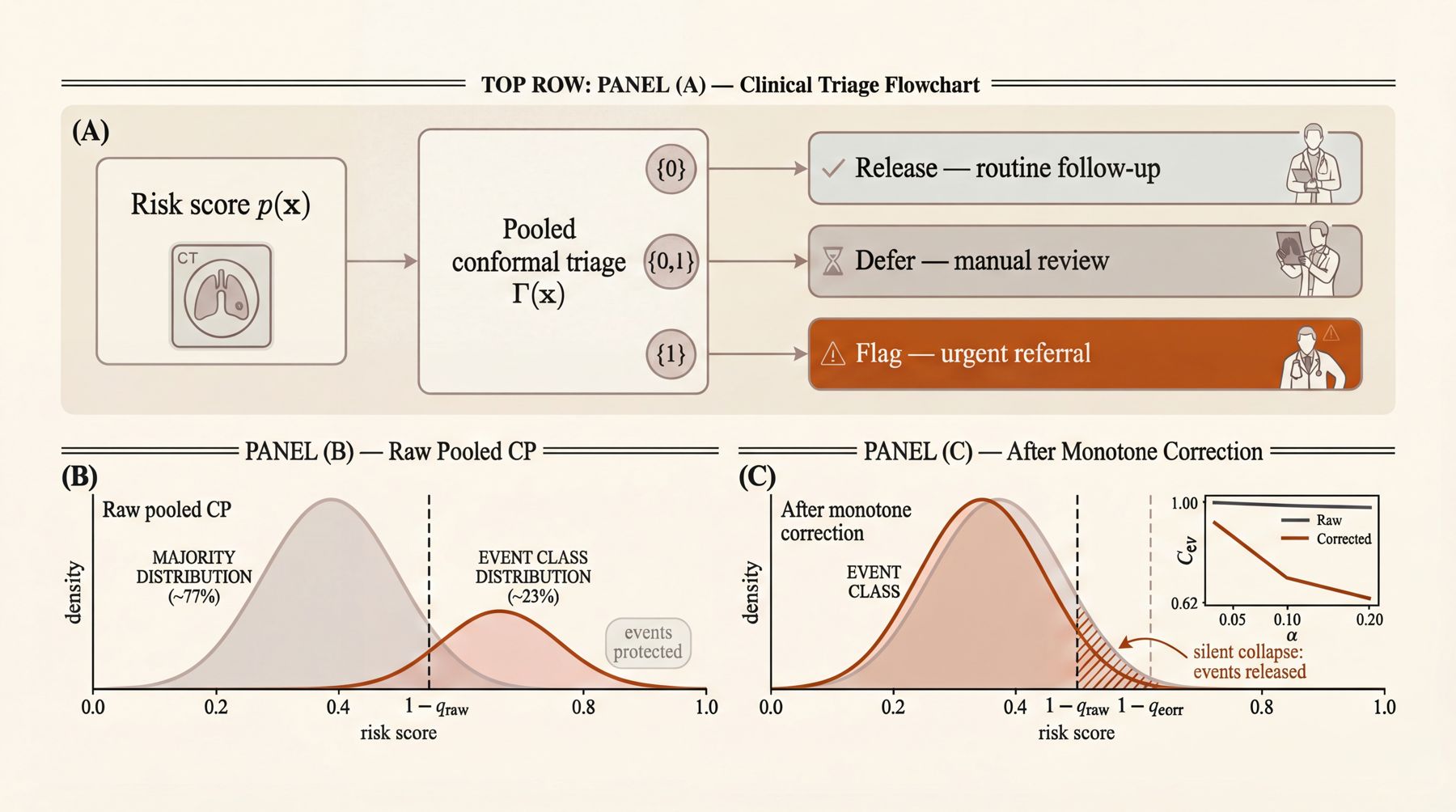}}
\end{figure}

Split conformal prediction (split CP), also known as inductive conformal prediction, augments a fixed predictive model with set-valued outputs whose marginal miscoverage is controlled in
finite samples under exchangeability \citep{vovk2005,shafer2008tutorial,angelopoulos2023conformal}.
Conformal prediction (CP) more broadly includes full and transductive variants that are not restricted to this post-hoc construction. A substantial line of work has refined how these sets are constructed for classification, from least-ambiguous set-valued classifiers that minimize set size at a target error level \citep{lei2014classification,sadinle2019leastambiguous} to adaptive prediction sets that align coverage with local difficulty \citep{romano2020}. In medical imaging, CP has been adapted into clinical triage pipelines that communicate per-patient uncertainty and route low-confidence cases to a human reviewer \citep{lu2022fairconformal,angelopoulos2024triage}. These pipelines share an operational semantics with classical selective classification, reject-option learning, and learning-to-defer \citep{chow1970,elyaniv2010,geifman2017selective,cortes2016learning,geifman2019selectivenet,mozannar2020defer}, in which the model is permitted to act on its own only when its output is reliable and otherwise hands the case to a human, and they share an evaluation paradigm with a parallel literature on out-of-distribution detection that flags inputs on which a prediction should not be trusted \citep{hendrycks2017baseline,liu2020energy,yang2024oodsurvey,li2025wrongquestions}. Across all of these lines, deployment quality is judged through aggregate summaries such as marginal coverage, set size, abstention rate, risk-coverage curves, and AUROC of an OOD score. None of these summaries directly answers the question a clinical operator must confront before deployment: among the patients the system clears without human review, are any of them true event cases who should not have been cleared, and how many slip through unattended?

To address prevalence shift in deployed classifiers, three families of methods have emerged: prior-probability and label-shift adjustments that reweight scores when class balance changes \citep{saerens2002,quionero2009dataset,lipton2018detecting,podkopaev2021}, classwise and conditional conformal methods that extend coverage beyond pooled marginal validity \citep{tibshirani2019covariate,romano2020,ding2023manyclasses,gibbs2025conditional}, and conformal risk control that targets user-specified risks beyond label inclusion \citep{bates2021distribution,angelopoulos2024conformalrisk}. A natural deployment pipeline combines a prevalence correction with pooled split conformal prediction. This pipeline implicitly assumes that a correction improving aggregate score alignment preserves the release behavior of the resulting rule. Under low target prevalence, the assumption should be assessed at the action level. The failure of interest is the unattended release of a true event case, which marginal label miscoverage does not capture. Deployment populations shift across sites, time periods, acquisition protocols, and disease prevalence, giving this audit practical importance \citep{moreno2012unifying,guo2017calibration,ovadia2019trust,wiens2019donoharm,feng2022monitoring,godau2023prevalence}. To investigate, we apply a labeled-pilot prevalence correction followed by pooled split conformal calibration on a retrospective NSCLC cohort, then trace each conformal set back to its deployment action and ask which event patients are released. Figure~\ref{fig:concept} summarizes the mechanism that motivates our audit and highlights two observations:

\begin{itemize}[leftmargin=*]
\item[\ding{182}] \textbf{\textit{Pooled marginal validity does not bound release-side event risk.}}
Pooled split conformal calibration sets a single threshold from all calibration patients, so in a low-prevalence target cohort that threshold is governed by non-event scores and marginal coverage can remain near its nominal level even when event-class coverage deteriorates well below the nominal target. This is not a violation of conformal validity, but a mismatch between the marginal guarantee and the clinical meaning of release.

\item[\ding{183}] \textbf{\textit{A useful-looking prevalence correction can still create unsafe releases.}}
A monotone prevalence correction may improve aggregate alignment and reduce review by producing more singleton low-risk outputs, yet because the pooled threshold remains anchored by the non-event majority, the corrected release boundary can sweep into a low-score region that still contains undetected true event cases. The resulting failure is action-level and visible only after asking which patients are actually released without review, not whether marginal coverage or workload looks acceptable.
\end{itemize}

These observations show that the failure is not caused by a missing conformal variant, but by how deployment decisions are calibrated, corrected, and evaluated. We therefore treat conformal triage as a release-side deployment system and audit the full decision path. Within each audit split, the protocol keeps prevalence correction, threshold calibration, and action-level evaluation on three disjoint
target subsets, so a label used to adjust scores or set thresholds is not reused to judge release decisions in that split. Once the decisions are fixed, the audit reports what the deployed system actually does, namely which event patients are released without review, how risky the released group is, how event coverage changes, and how much human review remains. This also clarifies the role of classwise calibration. If event releases decrease only because most patients return to review, the result is not evidence of useful low-review automation but evidence that the pilot data contain too few event labels to support release without review.

Our contributions are as follows:
\begin{itemize}
\item \textbf{Leakage-aware release audit.} Casting conformal triage as a clinically asymmetric instance of reject-option conformal classification, we define a release/flag/defer audit and report held-out release-side metrics on
$T$ under a disjoint $C_1/C_2/T$ protocol that isolates prevalence correction, threshold calibration, and evaluation.

\item \textbf{Finite-sample scarcity diagnostic.} We derive a closed-form fail-safe condition tying the size of the event calibration set to the target error rate. It predicts when classwise CP must default to high-review deferral and can no longer produce informative release, turning prevalence shift into an explicit label-allocation problem.

\item \textbf{Retrospective NSCLC evidence.} On a low-prevalence NSCLC cohort, the audit shows that a prevalence-corrected pooled rule purchases lower review burden by releasing true event patients without human review, while classwise calibration reduces event-positive releases only by returning nearly all patients to review, exposing calibration insufficiency without certifying useful low-review automation.
\end{itemize}

\section{Related Work}
\label{sec:related}

\paragraph{Conformal guarantees and set-valued classification.}
Conformal prediction provides finite-sample guarantees for set-valued prediction under exchangeability, while related work on classification with confidence and least-ambiguous set-valued classifiers studies how to trade label coverage against ambiguity \citep{vovk2005,shafer2008tutorial,angelopoulos2023conformal,lei2014classification,sadinle2019leastambiguous}. Classwise, label-conditional, subgroup-aware, and conditional conformal methods refine the target of validity beyond a single pooled marginal statement \citep{romano2020,ding2023manyclasses,gibbs2025conditional,lu2022fairconformal}. Conformal risk control further shifts attention from label inclusion to user-specified risks \citep{bates2021distribution,angelopoulos2024conformalrisk}. Our work is orthogonal to these guarantee types. We do not propose a new coverage guarantee or a new prediction-set construction. Instead, we study what happens after an existing conformal set is converted into a deployment action in downstream clinical use, where the clinically important failure is not merely label exclusion but the unattended release of a true event case. 

\paragraph{Selective prediction, deferral, and clinical review accounting.}
Selective classification, reject-option learning, and learning-to-defer study when a model should abstain or hand a case to an expert, typically through accuracy--coverage, risk--coverage, or expert-deferral trade-offs \citep{chow1970,elyaniv2010,geifman2017selective,cortes2016learning,geifman2019selectivenet,mozannar2020defer}. These formulations are closely related to conformal triage, but clinical triage has a more structured action space than a single abstention option: a case can be released, flagged for urgent review, or deferred for standard review \citep{angelopoulos2024triage}. This distinction matters for workload accounting in realistic clinical workflows. Counting high-risk flags as automated resolution would understate clinical workload, while counting only deferrals would miss the cost of urgent review. We therefore treat both flag and defer as human review and reserve ``release'' for cases cleared without immediate review. This convention lets the audit separate lower review burden from genuinely lower-risk automation.

\paragraph{Conformal classification with a reject option.}
A closely related line of work casts binary conformal prediction as a
classifier with a reject option: singleton sets are accepted as confident
predictions, while doubleton or empty sets are rejected and set aside
\citep{hallberg2025reject}. Our release/flag/defer rule is an instance of this
set-to-action logic, but with a clinically asymmetric action space: only the
negative singleton $\{0\}$ is released without review, the positive singleton
$\{1\}$ is flagged, and $\{0,1\}$ or $\emptyset$ is deferred. Relative to this
literature, our focus is not the accept/reject construction itself but the
release-side audit of that action under an observed prevalence change.

\paragraph{Prevalence shift, calibration, and target-label scarcity.}
Dataset shift and calibration failure are well-known barriers to reliable deployment, especially in clinical AI systems that move across sites, time periods, acquisition protocols, and disease prevalences \citep{moreno2012unifying,guo2017calibration,ovadia2019trust,wiens2019donoharm,feng2022monitoring,godau2023prevalence}. Prior-probability correction, label-shift adjustment, and distribution-shift-aware conformal methods address important aspects of changed class balance or changed test distributions \citep{saerens2002,quionero2009dataset,lipton2018detecting,podkopaev2021,tibshirani2019covariate}. These methods usually ask whether scores, weights, or prediction sets remain valid under a specified shift model. Our question is different and more operational: after a shift correction and conformal calibration have been fixed, does the target pilot contain enough event labels to support safe release decisions? This turns prevalence shift into a finite-label allocation problem. The same scarce target events, often with delayed outcome labels, must support correction, calibration, and held-out evaluation, so a low-review operating point can be unsupported even when standard marginal summaries look acceptable.

\section{Methodology: Release-Side Risk Audit}
\label{sec:methodology}

\subsection{Problem setting and overview}
\label{sec:problem_overview}

\begin{figure}[t]
  \centering
  \includegraphics[width=\textwidth]{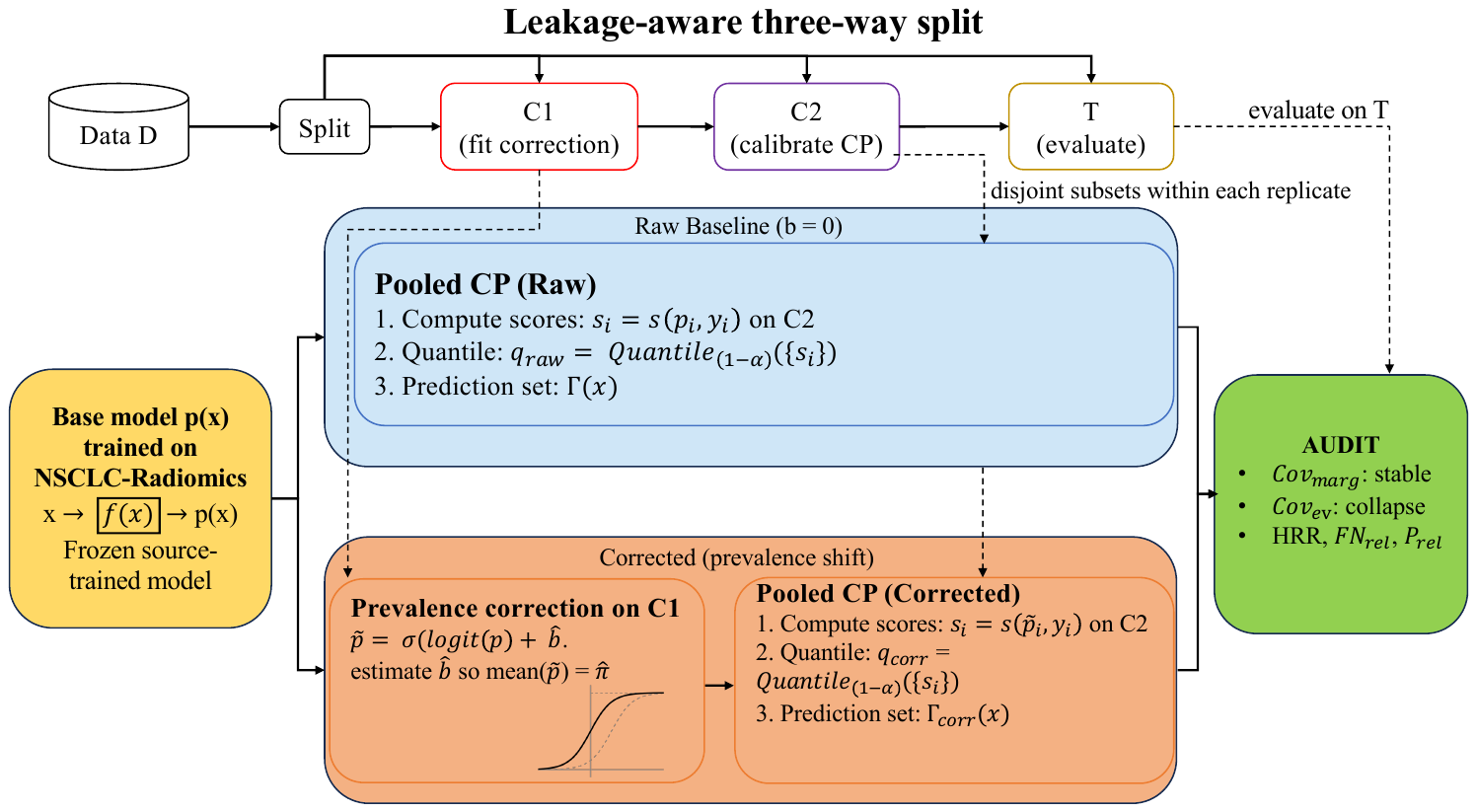}
  \caption{Leakage-aware pilot audit pipeline. The target cohort is split into three disjoint subsets. \(C_1\) estimates a labeled-pilot prevalence correction, \(C_2\) calibrates conformal thresholds, and \(T\) evaluates release-side and workload metrics. The raw and corrected branches use the same held-out evaluation set, so the audit asks whether prevalence correction changes release-side risk or only~reduces~review.}
  \label{fig:pipeline}
\end{figure}

The audit starts from an already trained risk model that outputs an event score for each target patient. We keep this model fixed and ask a deployment question: after the score is turned into conformal triage actions, can the system safely release low-risk patients in a lower-prevalence target cohort? The main obstacle is label accounting. Target labels are needed to estimate the prevalence correction, calibrate conformal thresholds, and evaluate whether released patients include true event cases. Reusing the same labels across these roles can make the audit overly optimistic, so Fig.~\ref{fig:pipeline} separates the target cohort into three non-overlapping parts with distinct purposes.

We use the term ``prevalence shift'' descriptively for the observed difference in event prevalence between the source and target analyzable cohorts. The audit assumes neither pure label shift nor invariance of $P(X\mid Y)$ across the two cohorts. Within each replicate the three subsets $C_1$, $C_2$, and $T$ are disjoint, so every target label supports at most one audit role. With the base model frozen and the correction fitted on $C_1$ and then held fixed, the split-conformal interpretation relies only on exchangeability between $C_2$ and $T$, and we make no claim of source-to-target exchangeability. Because every replicate reuses the same target cohort, the repeated splits report split sensitivity and do not constitute independent external validation.

The rest of this section follows the three operations in the pipeline. Section~\ref{sec:correction} explains how the labeled pilot subset fits a monotone prevalence correction, which is then frozen before calibration. Section~\ref{sec:metrics} defines how prediction sets become release, flag, or defer actions, and which release-side safety and workload metrics are measured on held-out patients. Section~\ref{sec:pooled_classwise} specifies the pooled and classwise conformal rules used in the raw and corrected branches. Evaluating both branches on the same held-out patients within each split makes review burden and release-side event risk paired rather than confounded comparisons.

\subsection{Labeled-pilot prevalence correction}
\label{sec:correction}

We study a retrospective labeled-pilot audit. The prevalence correction uses labeled outcomes in \(C_1\); it is therefore a delayed-outcome or retrospective pre-deployment audit, not an unlabeled real-time adaptation procedure. This distinction is important for clinical endpoints such as two-year event status. The correction is a monotone logit shift, that is, an intercept-only
recalibration in logit space,
\begin{equation}
  g_b(p)=\sigma(\operatorname{logit}(p)+b),\qquad
  \sigma(z)=\frac{1}{1+e^{-z}}.
\end{equation}
We fit the scalar $b$ by matching the corrected mean score on $C_1$
to the observed pilot prevalence, and we refer to the resulting map
$g_b$ as the prevalence-matching correction:
\begin{equation}
\frac{1}{|C_1|}\sum_{i\in C_1} g_b(\hat p_i)=\hat\pi_{C_1}
\triangleq \frac{1}{|C_1|}\sum_{i\in C_1}y_i .
\label{eq:b_match}
\end{equation}
Once \(b\) is fitted on the labeled pilot subset, it is frozen for the rest of the audit. Conformal thresholds are calibrated only on \(C_2\), and all audit metrics are computed only on \(T\).

Base-model fitting and post-hoc Platt calibration were completed using source-side data before any target-side audit labels were used. Thus, within each audit replicate, target labels enter only through the
three audit-specific roles in $C_1$, $C_2$, and $T$, ensuring that no target subject contributes labels to more than one target-side operation in that replicate.

\subsection{Conformal triage actions and release-side metrics}
\label{sec:metrics}

\paragraph{Terminology.}
We use $\HRR$ for the human-review rate, $\Cmarg$ and $\Cev$ for marginal
and event-class coverage, $\FNrel$ for the fraction of event-positive cases
released as $\{0\}$, and $\Prel$ for the event rate among released cases.

A conformal classifier produces a set $\Gamma(x)\subseteq\{0,1\}$. We map this set to deployment actions:
$$
a(x)=
\begin{cases}
\text{release}, & \Gamma(x)=\{0\},\\
\text{flag}, & \Gamma(x)=\{1\},\\
\text{defer}, & \Gamma(x)\in\bigl\{\{0,1\},\emptyset\bigr\}.
\end{cases}
$$
The empty set is treated as defer because it does not identify a release
decision. Let $r$, $f$, $d_{\mathrm{amb}}$, and $d_{\emptyset}$
denote the rates of the four possible set outcomes---the release singleton
$\{0\}$, the flag singleton $\{1\}$, the ambiguous doubleton $\{0,1\}$,
and the empty set $\emptyset$:
$$
r=\mathbb{P}(\Gamma(X)=\{0\}),\qquad
f=\mathbb{P}(\Gamma(X)=\{1\}),
$$
$$
d_{\mathrm{amb}}=\mathbb{P}(\Gamma(X)=\{0,1\}),\qquad
d_{\emptyset}=\mathbb{P}(\Gamma(X)=\emptyset).
$$
The doubleton and the empty set both map to the defer action, so the
human-review rate is
\begin{equation}
\HRR=1-r=f+d_{\mathrm{amb}}+d_{\emptyset}.
\end{equation}
Flagging is counted as review because it still consumes clinical attention; counting flags as automated resolution would create a workload loophole.

The standard conformal summary is marginal coverage,
\begin{equation}
\Cmarg=\mathbb{P}(Y\in\Gamma(X)).
\end{equation}
For release-side auditing, we also report event coverage,
\begin{equation}
\Cev=\mathbb{P}(1\in\Gamma(X)\mid Y=1),
\end{equation}
event-release risk,
\begin{equation}
\FNrel=\mathbb{P}(\Gamma(X)=\{0\}\mid Y=1),
\end{equation}
and the event prevalence among released cases,
\begin{equation}
\Prel=\mathbb{P}(Y=1\mid \Gamma(X)=\{0\}).
\end{equation}
We also report $\FNp$, the 95th percentile of $\FNrel$ across random
splits, and the release--review operating curve, which traces
$(\HRR(\alpha),\FNrel(\alpha))$ as $\alpha$ varies. Repeated splits
quantify split sensitivity, not independent external validation. Because
empty prediction sets are mapped to defer rather than release, for
event-positive cases
\begin{equation}
1-\Cev=\FNrel+\mathbb{P}(\Gamma(X)=\emptyset\mid Y=1).
\end{equation}
Event miscoverage can therefore arise either from an event-positive release
or from a conservative empty-set deferral.

\subsection{Conformal calibration rules}
\label{sec:pooled_classwise}

We use split conformal classification on the calibration subset $C_2$ after the score transformation, if any, has been fixed. For an event score \(p(x)\), define
\[
\hat p_1(x)=p(x), \qquad \hat p_0(x)=1-p(x),
\]
and use the nonconformity score
\begin{equation}
s(x,y)=1-\hat p_y(x), \qquad y\in\{0,1\}.
\label{eq:nonconformity}
\end{equation}
For the raw branch, \(p(x)=\hat p(x)\). For the corrected branch, \(p(x)=g_b(\hat p(x))\), where \(b\) has already been fitted on the labeled pilot subset and then frozen.

For a calibration multiset \(S=\{s_i\}_{i=1}^n\), let \(S_{(k)}\) denote the \(k\)-th smallest element and set
\begin{equation}
k_\alpha(n)=\left\lceil (n+1)(1-\alpha)\right\rceil .
\label{eq:conformal_index}
\end{equation}
If \(k_\alpha(n)\le n\), the conformal threshold is \(S_{(k_\alpha(n))}\). If \(k_\alpha(n)>n\), we use the conservative convention \(q=\infty\), so that the corresponding label is always included.

\paragraph{Pooled calibration.}
Pooled split conformal prediction forms one calibration multiset from all labeled calibration patients,
\[
S_{\mathrm{pool}}=\{s(x_i,y_i): i\in C_2\},
\]
and computes a single threshold
\[
\hat q_{\mathrm{pool}} = S_{\mathrm{pool},(k_\alpha(|C_2|))}.
\]
The prediction set for a held-out patient is
\begin{equation}
\Gamma_{\mathrm{pool}}(x)=
\{y\in\{0,1\}: s(x,y)\le \hat q_{\mathrm{pool}}\}.
\label{eq:pooled_set}
\end{equation}
This rule targets marginal coverage on the target distribution because event and non-event calibration scores share a single global threshold.

\paragraph{Classwise calibration.}
Classwise split conformal prediction calibrates a separate threshold for each label. For each $y\in\{0,1\}$, define
$$
S_y=\{s(x_i,y_i): i\in C_2,\ y_i=y\}, \qquad n_y=|S_y|.
$$
The classwise threshold is
$$
\hat q_y =
\begin{cases}
S_{y,(k_\alpha(n_y))}, & k_\alpha(n_y)\le n_y,\\
\infty, & k_\alpha(n_y)>n_y.
\end{cases}
$$
The prediction set is
\begin{equation}
\Gamma_{\mathrm{cw}}(x)=
\{y\in\{0,1\}: s(x,y)\le \hat q_y\}.
\label{eq:classwise_set}
\end{equation}
The infinite-threshold convention is important in small event-calibration samples: if the event threshold is infinite, the event label is always included, so a pure low-risk release $\{0\}$ is impossible. Section~\ref{sec:diagnostics} makes this precise by giving the exact event-count condition that triggers this fail-safe regime. In this paper, classwise calibration is therefore used as a scarcity diagnostic, not as a newly proposed conformal algorithm.

\section{Finite-Sample Audit Mechanism}
\label{sec:diagnostics}

This section records the finite-sample mechanisms specific to release-side triage that the audit will test empirically in Sec.~\ref{sec:nsclc}. These observations do not define a new conformal algorithm. They explain why ordinary coverage and workload summaries can be insufficient for release-side triage safety under low event prevalence.

\paragraph{Obs.~\ding{182} Marginal coverage gives only a weak event-safety floor.}
Let \(\pi=\mathbb{P}(Y=1)\). If \(\mathbb{P}(Y\in\Gamma(X))\ge 1-\alpha\), then
\begin{equation}
\Cev\ge \max\{0,1-\alpha/\pi\}.
\label{eq:floor}
\end{equation}
This follows directly from total probability:
\[
1-\alpha\le \pi\Cev+(1-\pi)\mathbb{P}(0\in\Gamma(X)\mid Y=0)\le \pi\Cev+(1-\pi).
\]
For the Radiogenomics target prevalence \(\pi=0.228\) and \(\alpha=0.10\), the bound is only \(0.561\). Thus a marginally valid triage system can still have poor event-class protection. This is the basic reason that release-side event risk must be reported separately from \(\Cmarg\).

\paragraph{Obs.~\ding{183} Monotone score correction leaves classwise ranks unchanged but can move a pooled boundary.}
Strictly increasing score remappings preserve within-class ranks. Therefore, for a fixed calibration split and under the usual exchangeability assumptions, classwise conformal thresholds produce the same label-inclusion decisions before and after a monotone remapping. Pooled conformal prediction does not have this invariance under monotone remapping because its single global quantile mixes event and non-event calibration scores. This distinction explains why a prevalence correction can leave classwise decisions stable while changing the pooled release boundary.

\paragraph{Obs.~\ding{184} Event-label scarcity forces classwise calibration into fail-safe behavior.}
Classwise conformal prediction estimates a \((1-\alpha)\)-quantile within each class. If \(n_y\) calibration points are available for class \(y\), the conformal index is \(k=\lceil(n_y+1)(1-\alpha)\rceil\). When \(k>n_y\), no finite order statistic is available, and the conservative convention forces a fail-safe output. This occurs if and only if
\begin{equation}
  n_y \le \lceil 1/\alpha\rceil -2.
\label{eq:fsrl_condition}
\end{equation}
At \(\alpha=0.10\), the event-class fail-safe triggers when \(n_1\le 8\). Thus classwise CP can reduce event releases not by enabling useful low-review automation, but by making release unavailable for most cases. This converts event-label scarcity into an explicit deployment limitation that cannot be removed by tuning the conformal threshold alone. Table~\ref{tab:fsrl_sizing} translates this condition into the minimum calibration size required to keep the fail-safe probability below a target risk \(\delta\), under both an i.i.d.\ deployment model (binomial) and the finite NSCLC audit cohort (hypergeometric).

\begin{table}[t]
\centering
\caption{Calibration sizing at $\alpha=0.10$ for the NSCLC target prevalence $\pi=0.228$ and finite audit cohort $N=123,K=28$. The event-class fail-safe triggers if $n_1\le 8$. The table reports the minimum calibration size $n_{\mathrm{cal}}$ needed to keep $\mathbb{P}(n_1\le 8)\le\delta$.}
\label{tab:fsrl_sizing}
\setlength{\tabcolsep}{7pt}
\begin{tabular}{lcc}
\toprule
Max fail-safe risk $\delta$ & Deploy model: Binomial & Finite audit: Hypergeometric\\
\midrule
$\le 50\%$ & $n_{\mathrm{cal}}\ge 38$ & $n_{\mathrm{cal}}\ge 38$\\
$\le 25\%$ & $n_{\mathrm{cal}}\ge 46$ & $n_{\mathrm{cal}}\ge 45$\\
\textbf{$\le 10\%$ default} & \textbf{$n_{\mathrm{cal}}\ge 55$} & \textbf{$n_{\mathrm{cal}}\ge 51$}\\
$\le 5\%$ & $n_{\mathrm{cal}}\ge 60$ & $n_{\mathrm{cal}}\ge 54$\\
$\le 1\%$ & $n_{\mathrm{cal}}\ge 72$ & $n_{\mathrm{cal}}\ge 61$\\
\bottomrule
\end{tabular}
\end{table}

\section{Retrospective NSCLC Pilot Audit}
\label{sec:nsclc}

We use the NSCLC audit to answer four questions. \textbf{RQ1} asks whether pooled prevalence correction can reduce review burden while increasing event releases. \textbf{RQ2} asks whether classwise conformal prediction reduces event-positive release through selective automation or mainly through deferral. \textbf{RQ3} asks whether the high-review behavior is explained by finite-sample event-calibration scarcity. \textbf{RQ4} asks whether secondary baselines and operating curves at other risk levels change the release-side interpretation of the pooled-vs-classwise comparison.

\subsection{Experimental setup}
\label{sec:setup_exp}

We evaluate two public TCIA NSCLC cohorts: NSCLC-Radiomics as the source cohort
and NSCLC-Radiogenomics as the lower-prevalence target cohort. The binary
endpoint $Y$ is 24-month mortality, with $Y=1$ for death within 24 months
and $Y=0$ for patients event-free through 24 months. We filter the original
211 Radiogenomics subjects for feature completeness and endpoint observability,
the latter excluding patients censored before 24 months without an observed
death, which leaves $N=123$ analyzable patients with $K=28$ events and gives
$\pi=0.228$. The audit is therefore a complete-case binary analysis that does
not model censoring. Its target cohort is a convenience analyzable subset
assembled retrospectively, so it does not represent a prospectively enrolled
deployment population with complete outcome ascertainment at the 24-month
horizon.

Base models are trained on NSCLC-Radiomics and applied to Radiogenomics; we
audit a Clinical model and a Radiomics model using frozen source-trained
predictions. The target cohort is repeatedly split into $C_1=31$, $C_2=31$,
and $T=61$, and unless otherwise stated the primary audit uses 200
unstratified random splits. Because only 28 events are available, the reported
mean$\pm$std reflects split sensitivity, not independent validation: the audit
protocol and event-count cutoff are general, but the observed magnitudes of
$\HRR$, event coverage, and event-release risk are model- and cohort-specific
and do not constitute clinical validation or support unattended deployment.

\subsection{Pooled correction lowers review but increases event release (RQ1)}
\label{sec:rq1}

\begin{figure}[t]
  \centering
  \includegraphics[width=0.95\textwidth]{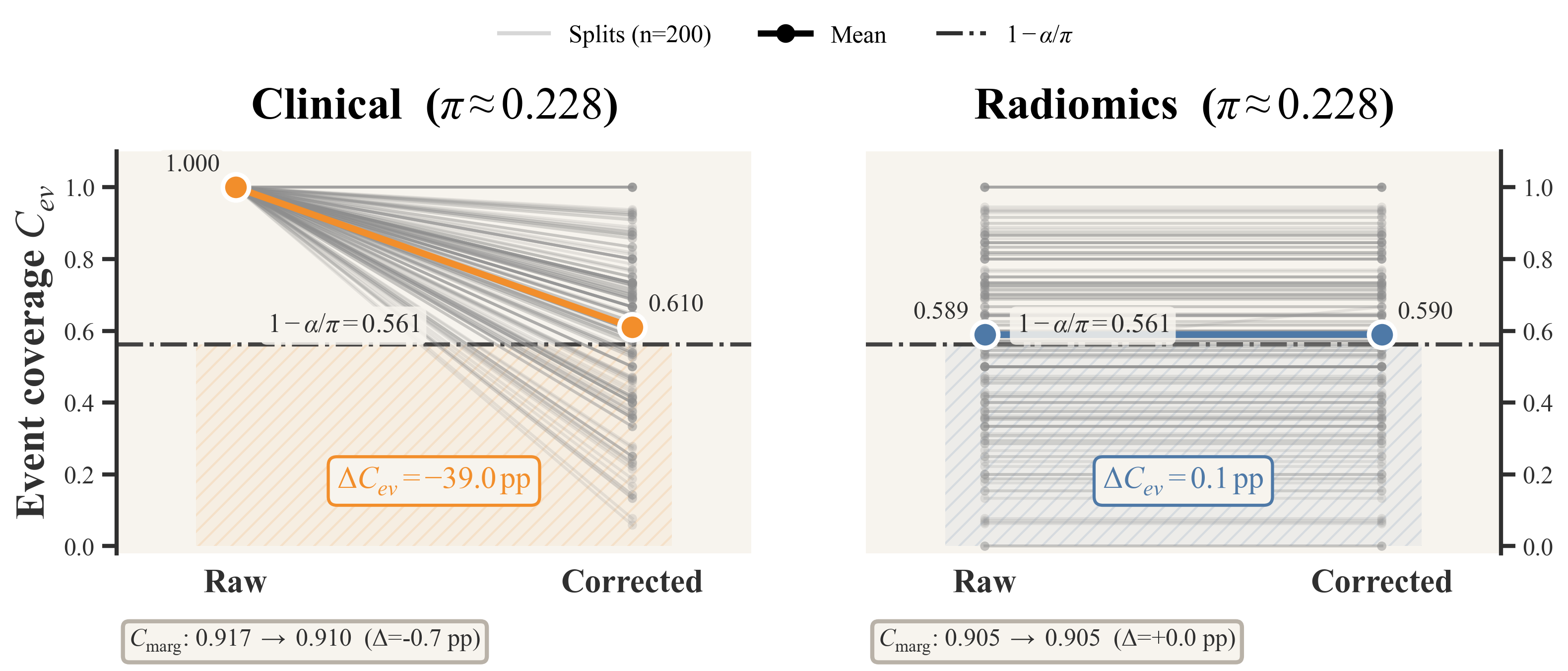}
  \caption{Event-coverage collapse under pooled prevalence correction at \(\alpha=0.10\). Left: for the Clinical model, event coverage drops from 1.000 to 0.610 while marginal coverage stays near nominal. Right: for the Radiomics model, raw and corrected results nearly overlap, showing that the audit identifies when a correction shifts the pooled boundary enough to create release-side failure.}
  \label{fig:collapse}
\end{figure}

Figure~\ref{fig:collapse} summarizes the pooled-threshold failure at \(\alpha=0.10\). For the Clinical model, pooled correction reduces the human-review rate from 1.000 to 0.425, but event coverage drops from 1.000 to 0.610 and \(\FNrel\) rises to 0.390. Marginal coverage changes only slightly, from 0.917 to 0.910. Thus the workload improvement is visible, while the release-side failure is hidden unless event-level action metrics are reported.

\paragraph{Obs.~\ding{185} Marginal coverage hides release-side failure.}
The Clinical branch shows the central failure mode: pooled correction reduces \(\HRR\) but increases event releases. This is consistent with Obs.~\ding{182}: at target prevalence \(\pi=0.228\) and \(\alpha=0.10\), marginal \(90\%\) coverage only guarantees \(C_{\mathrm{ev}}\ge 0.561\). The all-review behavior of Pooled-Raw is also informative. At this \(\alpha\) and calibration size, the raw score distribution does not support singleton low-risk release under the pooled threshold. The correction creates releasable singletons, but the audit reveals that these releases include true event cases.

The Radiomics branch serves as an internal negative control: raw and corrected results nearly overlap. This shows that pooled correction does not universally increase event-positive releases; the audit detects
model-dependent changes in the pooled release boundary.

\subsection{Classwise CP reduces event release mainly by deferral (RQ2)}
\label{sec:rq2}

Table~\ref{tab:main} reports the \(\alpha=0.10\) comparison. The first block tests the paper's main mechanism: pooled correction can reduce review while increasing event release, whereas classwise CP suppresses event release by becoming highly conservative. The second block reports risk-controlling prediction sets (RCPS)
and Label-Shift CP as diagnostic baselines; they contextualize conservativeness and shift-assumption sensitivity, but are not used to claim method superiority.

\begin{table}[t]
\centering
\caption{Release-side audit at $\alpha=0.10$ over 200 random target splits. Mean$\pm$std summarizes split sensitivity, not external validation. The first block tests the paper's main mechanism; the second reports diagnostic baselines for context only. The metrics form a release--review trade-off and are read
jointly; no single column defines a winning method.}
\label{tab:main}
\setlength{\tabcolsep}{2.5pt}
\renewcommand{\arraystretch}{1.04}
\footnotesize
\begin{adjustbox}{width=\textwidth, center}
\begin{tabular}{lcccccccc}
\toprule
& \multicolumn{4}{c}{Clinical Base Model} 
& \multicolumn{4}{c}{Radiomics Base Model} \\
\cmidrule(lr){2-5}\cmidrule(lr){6-9}
Method
& \(\HRR\)
& \(\FNrel\)
& \(Q_{0.95}(\FNrel)\)
& \(\Prel\)
& \(\HRR\)
& \(\FNrel\)
& \(Q_{0.95}(\FNrel)\)
& \(\Prel\) \\
\midrule
\multicolumn{9}{l}{\emph{Primary audit comparison}}\\
Pooled-Raw
& \(1.000\) & \(0.000\) & \(0.000\) & ---
& \(0.478\sd{.217}\) & \(0.411\sd{.235}\) & \(0.825^{\dagger}\) & \(0.169\) \\
Pooled-Corr
& \(0.425\sd{.171}\) & \(0.390\sd{.215}\) & \(0.770^{\dagger}\) & \(0.148\)
& \(0.478\sd{.217}\) & \(0.410\sd{.235}\) & \(0.825^{\dagger}\) & \(0.169\) \\
Classwise
& \(0.956\sd{.106}\) & \(0.017\sd{.051}\) & \(0.167\) & \(0.076\)
& \(0.980\sd{.049}\) & \(0.019\sd{.059}\) & \(0.167\) & \(0.168\) \\
\midrule
\multicolumn{9}{l}{\emph{Secondary diagnostic baselines}}\\
RCPS
& \(1.000\) & \(0.000\) & \(0.000\) & ---
& \(1.000\) & \(0.000\) & \(0.000\) & --- \\
Label-Shift CP
& \(1.000\) & \(0.000\) & \(0.000\) & ---
& \(0.080\sd{.067}\) & \(0.845\sd{.109}\) & \(1.000^{\dagger}\) & \(0.208\) \\
\bottomrule
\end{tabular}
\end{adjustbox}

\vspace{2pt}
\begin{minipage}{0.98\textwidth}
\footnotesize
$\dagger$ marks $\FNp \ge 0.75$; it flags high split-tail event-release risk and does not denote statistical significance. ``---'' indicates no cases were released, making $\Prel$ undefined. $\Prel$ is the mean of split-specific released-group event rates over splits with at least one
released case.
\end{minipage}
\end{table}

\paragraph{Obs.~\ding{186} Classwise event-release reduction is not free automation.}
For the Clinical model, classwise CP reduces \(\FNrel\) from $0.390$ to $0.017$, but its \(\HRR\) is $0.956$. This should not be interpreted as successful low-review deployment. It indicates that event-release risk is reduced primarily by returning patients to review. The Radiomics model shows the same qualitative pattern: classwise CP keeps \(\FNrel\) low, but only under very high review.

\subsection{Calibration scarcity explains the high-review regime (RQ3)}
\label{sec:rq3}

Table~\ref{tab:ncal_sensitivity} traces calibration-size sensitivity for the Clinical model at $\alpha=0.10$. As $C_2$ grows, the probability that $n_1(C_2)\le 8$ decreases, while the held-out set $T$ becomes smaller. At the primary allocation $|C_2|=31$, the expected event count is $7.1$, below the nine event labels required for a finite event-class threshold. For $|C_2|\ge 40$, the expected count exceeds this cutoff, although unstratified splits still enter the infinite-threshold regime with probability $0.396$ at $|C_2|=40$ and $0.220$ at $|C_2|=45$. The result is therefore a target-label allocation trade-off, not an unavoidable regime under every allocation.

\begin{table}[t]
\centering
\caption{Calibration-size sensitivity for the Clinical model at \(\alpha=0.10\), computed over 1000 random splits for each allocation. Increasing \(C_2\) sharply reduces event-class fail-safe probability, but also leaves fewer held-out evaluation subjects in $T$ within each audit replicate.}
\label{tab:ncal_sensitivity}
\setlength{\tabcolsep}{3.5pt}
\renewcommand{\arraystretch}{1.05}
\footnotesize
\begin{adjustbox}{width=\textwidth, center}
\begin{tabular}{cccccccc}
\toprule
\(|C_2|\) & \(|T|\) & \(E[n_1(C_2)]\) & \(\mathbb{P}(n_1(C_2)\le 8)\) & Pooled-Corr \(\HRR\) & Pooled-Corr \(\FNrel\) & Classwise \(\HRR\) & Classwise \(\FNrel\)\\
\midrule
20 & 72 & 4.6 & 0.986 & 0.427 & 0.401 & 0.997 & 0.001 \\
31 & 61 & 7.1 & 0.766 & 0.420 & 0.403 & 0.952 & 0.020 \\
40 & 52 & 9.1 & 0.396 & 0.403 & 0.420 & 0.889 & 0.048 \\
45 & 47 & 10.2 & 0.220 & 0.446 & 0.374 & 0.860 & 0.061 \\
51 & 41 & 11.6 & 0.086 & 0.409 & 0.414 & 0.849 & 0.068 \\
55 & 37 & 12.5 & 0.040 & 0.438 & 0.373 & 0.845 & 0.067 \\
61 & 31 & 13.9 & 0.010 & 0.402 & 0.417 & 0.851 & 0.066 \\
\bottomrule
\end{tabular}
\end{adjustbox}
\end{table}

\paragraph{Obs.~\ding{187} The audit exposes a target-label allocation bottleneck.}
Across the calibration sizes explored, classwise $\HRR$ declines as $C_2$
grows yet remains far from a low-review regime, while classwise $\FNrel$ stays
well below the pooled-corrected $\FNrel$. What limits performance is therefore
not which threshold the rule selects but how few target events are available to
estimate one: the same scarce events must support correction fitting,
event-threshold calibration, and held-out release-risk evaluation. Within this
finite-event budget, reducing calibration scarcity necessarily leaves fewer
labels for held-out evaluation.

\subsection{Secondary baselines, operating curves, and split sensitivity (RQ4)}
\label{sec:rq4}

The diagnostic baselines do not change the main interpretation. RCPS is conservative and defers all cases at this calibration size, reinforcing the calibration-scarcity conclusion. Label-Shift CP behaves poorly on the Radiomics branch under its shift assumptions; we therefore treat it as an assumption-sensitivity diagnostic rather than as a primary competitor.

\begin{figure}[t]
  \centering
  \includegraphics[width=0.9\textwidth]{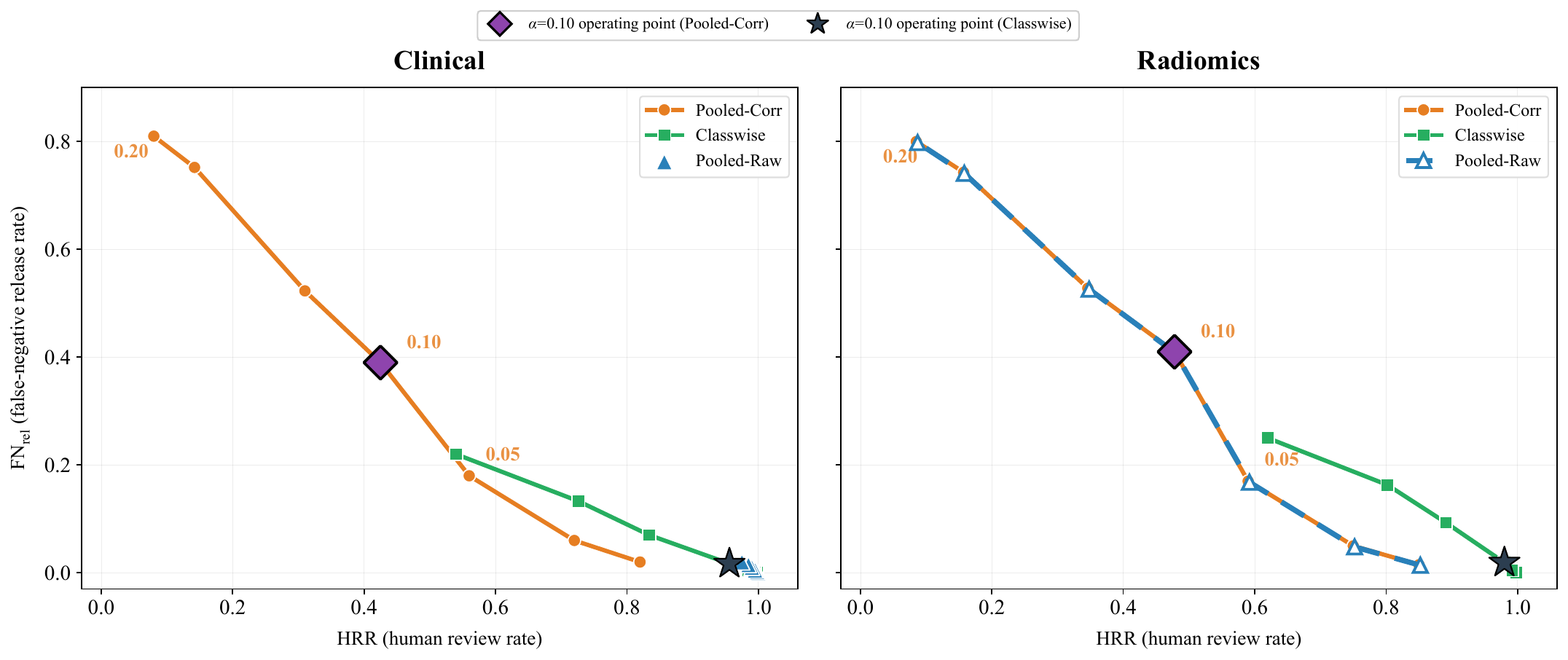}
\caption{Release--review operating curves over
$\alpha\in\{0.01,0.02,0.05,0.10,0.15,0.20,0.30\}$.
Labels on Pooled-Corr show $\alpha$. In the Clinical model, pooled
correction lowers review at the cost of higher event-release risk,
whereas classwise CP lowers this risk only at high review. Pooled-Raw
remains near full review for Clinical and overlaps Pooled-Corr for
Radiomics; the curves are descriptive and do not define a clinical
threshold for deployment decisions.}
  \label{fig:ctoc}
\end{figure}

Figure~\ref{fig:ctoc} plots the release--review operating curve across \(\alpha\in\{0.01,\ldots,0.30\}\). For the Clinical model, correction shifts the operating point toward lower review and higher event release. Changing \(\alpha\) moves along the operating curve; it does not remove the underlying release-side trade-off.

\paragraph{Obs.~\ding{188} Operating curves confirm the same release-review trade-off.}
Across $\alpha$, pooled correction moves the Clinical system toward lower review
by accepting higher event-release risk, while classwise CP moves toward lower
event release at high review. Secondary baselines do not overturn this reading:
RCPS collapses to full review, and Label-Shift CP is sensitive to its shift
assumptions. What the audit exposes is therefore a property of the entire
operating curve, not an artifact of any single risk level $\alpha$.

\section{Conclusion}
\label{sec:conclusion}

We introduced a leakage-aware, action-level audit of conformal triage under an
observed change in event prevalence. Within each replicate, prevalence
correction, conformal calibration, and held-out evaluation occupy disjoint
patient subsets; every prediction set is read as a release, flag, or defer
action rather than as a coverage statistic; and pooled and classwise rules are
compared on the same held-out patients, so review burden and release-side event
risk become paired quantities. On a retrospective NSCLC cohort this exposed two
failures that marginal coverage cannot see: a prevalence-corrected pooled rule
lowered the review rate while releasing true event cases, and classwise
calibration suppressed those releases mainly by returning cases to review. At a
fixed $\alpha$, the latter is a fail-safe response to too few event labels,
governed by the event count that prevalence and label allocation induce rather
than by threshold tuning. The evidence is a single retrospective, complete-case
cohort whose binary endpoint does not model censoring, so this is a mechanism
audit rather than clinical validation; release-side risk and calibration-scarcity
diagnostics nonetheless belong alongside marginal coverage and workload in any
deployment assessment of conformal triage, particularly when the target event is
rare and target labels are scarce.

\newpage
\bibliography{references}

\clearpage
\appendix
\input{appendix}

\end{document}

%% file: appendix.tex
\section{Operational Label-Use Accounting}
\label{app:operational}

Within each audit replicate, target labels used to estimate the
prevalence correction are not reused to calibrate conformal
thresholds, and labels used for either step are not reused to
evaluate release decisions.

\begin{table}[htbp]
\centering
\caption{Label-use ledger for the retrospective audit. Within each audit replicate, no target subject label is used for more than one target-side operation.}
\label{tab:app_label_use}
\setlength{\tabcolsep}{4pt}
\renewcommand{\arraystretch}{1.08}
\footnotesize
\begin{adjustbox}{width=\textwidth,center}
\begin{tabular}{p{0.23\textwidth}p{0.25\textwidth}p{0.25\textwidth}p{0.18\textwidth}}
\toprule
Operation & Data used & Target labels used? & Frozen before evaluating \(T\)? \\
\midrule
Base-model training & Source cohort only & No target labels & Yes \\
Post-hoc Platt calibration & Source-side validation split only & No target labels & Yes \\
Prevalence correction & \(C_1\) & Yes, \(y_i\) for \(i\in C_1\) & Yes \\
Conformal thresholding & \(C_2\) & Yes, \(y_i\) for \(i\in C_2\) & Yes \\
Audit metrics & \(T\) & Yes, \(y_i\) for \(i\in T\) & Evaluation only \\
\bottomrule
\end{tabular}
\end{adjustbox}
\end{table}

The correction is a labeled-pilot or retrospective pre-deployment operation. It is not an unlabeled real-time adaptation method. For a delayed endpoint such as a 24-month mortality endpoint, \(C_1\) labels are available only after outcome ascertainment or in retrospective audit. Base-model fitting and source-side Platt calibration are completed before any target-side labels are used.

\section{Finite-Sample Derivations}
\label{app:derivations}

The main text states the finite-sample mechanisms used by the audit. We include the derivations here to make clear that they explain the audit behavior rather than define a new conformal algorithm.

\paragraph{Marginal coverage and event coverage.}
Let \(\pi=\mathbb{P}(Y=1)\). If \(\mathbb{P}(Y\in\Gamma(X))\ge1-\alpha\), then
\[
\mathbb{P}(Y\in\Gamma(X))
=
\pi\,\mathbb{P}(1\in\Gamma(X)\mid Y=1)
+
(1-\pi)\,\mathbb{P}(0\in\Gamma(X)\mid Y=0).
\]
Since the second term is at most \(1-\pi\),
\[
1-\alpha \le \pi\Cev+(1-\pi),
\qquad
\Cev\ge \max\left\{0,1-\frac{\alpha}{\pi}\right\}.
\]
For Radiogenomics, \(\pi=28/123=0.228\) and \(\alpha=0.10\), so marginal \(90\%\) coverage only implies \(\Cev\ge0.561\). Thus substantial event-class failure is compatible with marginal validity.

\paragraph{Monotone remapping and classwise ranks.}
Let \(f:(0,1)\to(0,1)\) be strictly increasing. For label \(1\), \(s(x,1)=1-\hat p(x)\) becomes \(s_f(x,1)=1-f(\hat p(x))\), which is a strictly increasing transform of \(s(x,1)\). For label \(0\), \(s(x,0)=\hat p(x)\) becomes \(s_f(x,0)=f(\hat p(x))\), again preserving within-class order. Classwise conformal calibration depends only on within-class order statistics, so classwise inclusion decisions are invariant to a strictly monotone score remapping on a fixed split. Pooled calibration does not share this invariance because it mixes event and non-event scores in one global quantile.

\paragraph{Finite event-class threshold condition.}
For class \(y\) with \(n_y\) calibration samples, split conformal prediction uses
\[
k_y=\left\lceil (n_y+1)(1-\alpha)\right\rceil.
\]
A finite order statistic exists only when \(k_y\le n_y\). Thus the fail-safe regime occurs when
\[
n_y\le \left\lceil\frac{1}{\alpha}\right\rceil-2.
\]
At \(\alpha=0.10\), this gives \(n_y\le8\). For the finite target cohort with \(N=123\), \(K=28\), and \(n_{\mathrm{cal}}=31\),
\[
\mathbb{P}(n_1\le 8)=
\sum_{j=0}^{8}
\frac{\binom{28}{j}\binom{95}{31-j}}{\binom{123}{31}}
\approx 0.766.
\]
This explains why classwise CP often enters a high-review regime under the reported split design.

\section{NSCLC Cohort and Split Accounting}
\label{app:nsclc_details}

The target Radiogenomics cohort starts from 211 subjects. We exclude
67 subjects with incomplete imaging features or clinical covariates
and 21 subjects censored before 24 months without an observed death.
The final analyzable target cohort contains $N=123$ subjects,
including $K=28$ deaths observed within 24 months. Subjects who died
within 24 months are labeled $Y=1$, and subjects known to be
event-free through 24 months are labeled $Y=0$. The audit therefore
estimates release-side risk for this analyzable subset rather than
for the original cohort.

\begin{table}[htbp]
\centering
\caption{Cohort and split accounting in the retrospective NSCLC audit.}
\label{tab:app_cohort_split}
\setlength{\tabcolsep}{4pt}
\renewcommand{\arraystretch}{1.08}
\footnotesize
\begin{adjustbox}{width=\textwidth,center}
\begin{tabular}{p{0.26\textwidth}p{0.20\textwidth}p{0.24\textwidth}p{0.22\textwidth}}
\toprule
Item & Role & Size / prevalence & Use \\
\midrule
NSCLC-Radiomics / LUNG1 & Source cohort & \(N=420,\ \pi=0.598\) & Base-model training \\
NSCLC-Radiogenomics & Target cohort & \(N=123,\ K=28,\ \pi=0.228\) & \(C_1/C_2/T\) audit \\
Correction subset & Target split & \(|C_1|=31\) & Fit prevalence correction \\
Calibration subset & Target split & \(|C_2|=31\) & Fit conformal thresholds \\
Evaluation subset & Target split & \(|T|=61\) & Report release-side metrics \\
\bottomrule
\end{tabular}
\end{adjustbox}
\end{table}

The full-cohort target AUCs are 0.689 for the Clinical score and
0.618 for the Radiomics score. These values are reported only as
post-hoc descriptions of the frozen target scores and were not used
for correction fitting, conformal calibration, selection of
$\alpha$, or operating-point selection.

The primary audit uses 200 random unstratified splits of the same \(N=123\) target cohort. These repeated splits quantify sensitivity to target split allocation; they do not constitute independent external validation cohorts. Under the reported split design, \(C_2\) contains an expected \(31\cdot28/123=7.06\) event cases, with median 7, while the fail-safe threshold at \(\alpha=0.10\) is \(n_1\le8\). This finite event-count scarcity explains the high-review behavior of classwise CP in the main results.

\section{Supplementary Checks and Reproducibility}
\label{app:secondary_reproducibility}

Secondary baselines serve as diagnostic references, not as primary competitors.
Risk-controlling prediction sets (RCPS) defer all cases at this calibration
size, consistent with the calibration-scarcity conclusion. Label-Shift CP is
treated as an assumption-sensitivity diagnostic because its performance depends
on the shift model.

PathMNIST is included only as a limited non-clinical sensitivity
check. It is not a patient-level deployment cohort, and we do not treat
it as a second clinical validation dataset. At $\alpha=0.10$, event
coverage decreases from $0.715$ before correction to $0.624$ after
correction. Because we report only event coverage here without the
complete $C_1/C_2/T$ release-side audit, this result is interpreted
only as a consistency check under a different score distribution, and it
provides no evidence of cross-domain or clinical generalization. All
clinical conclusions are based on the retrospective NSCLC audit.

The audit implementation records the fixed random split generator,
all $C_1/C_2/T$ indices, fitted correction parameters, conformal
thresholds, per-subject triage decisions on $T$, and scripts for
the reported tables and release--review operating curves.